%% file: emnlp2022.tex
\pdfoutput=1

\documentclass[11pt]{article}

\usepackage{EMNLP2022}

\usepackage{times}
\usepackage{latexsym}

\usepackage[T1]{fontenc}

\usepackage[utf8]{inputenc}

\usepackage{microtype}
\usepackage{todonotes}

\usepackage{times}
\usepackage{helvet}  
\usepackage{courier}  
\usepackage{latexsym}
\usepackage{amsfonts}
\usepackage{amssymb}
\usepackage{amsmath,amsthm}
{
      \theoremstyle{plain}
      \newtheorem{objective}{Objective}
  }
\usepackage{dsfont}
\usepackage{bm}
\usepackage{booktabs}
\usepackage{multirow}
\usepackage{graphicx}  
\usepackage{comment}
\usepackage[normalem]{ulem}
\usepackage{xcolor}
\usepackage{makecell}
\usepackage{algorithm}
\usepackage{algpseudocode}
\usepackage{mathtools}
\usepackage{enumitem}
\usepackage{url}
\usepackage{xspace}
\usepackage{bbm}
\usepackage{inconsolata}

\DeclareMathOperator{\D}{\mathcal{D}}
\DeclareMathOperator{\T}{\mathcal{T}}

\DeclareMathOperator{\LL}{\mathcal{L}}
\DeclareMathOperator{\DD}{\mathbb{D}}
\DeclareMathOperator{\SSS}{\Omega}
\DeclareMathOperator{\mS}{\mathcal{S}}
\DeclareMathOperator{\R}{\mathbb{R}}

\newcommand{\mprompt}{\textsc{\small{ModularPrompt}}\xspace}
\definecolor{darkBlue}{RGB}{47, 85, 152}
\definecolor{lightPink}{RGB}{237, 219, 221}

\usepackage[capitalize,noabbrev]{cleveref}
\crefformat{section}{\S#2#1#3} 
\crefformat{subsection}{\S#2#1#3}
\crefformat{subsubsection}{\S#2#1#3}

\newcommand{\red}[1]{\textcolor{red}{#1}}

\newcommand{\darkBlue}[1]{\textcolor{darkBlue}{#1}}

\title{Learning Label Modular Prompts for Text Classification in the Wild}

\author{Hailin Chen$^\clubsuit$$^\spadesuit$,
Amrita Saha$^\spadesuit$,
Shafiq Joty$^\clubsuit$$^\spadesuit$,
Steven C.H. HOI$^\spadesuit$ \\
$^\clubsuit$ Nanyang Technological University, Singapore\\
$^\spadesuit$ Salesforce Research\\
\texttt{\{hailin001, srjoty\}@ntu.edu.sg}\\
\texttt{\{amrita.saha, shoi\}@salesforce.com}
}

\begin{document}
\maketitle
\begin{abstract}

Machine learning models usually assume i.i.d data during training and testing, but data and tasks  in real world often change over time. To emulate the transient nature of real world, we propose a challenging but practical task: text classification \textit{in-the-wild}, which introduces different non-stationary training/testing stages. Decomposing a complex task into modular components can enable robust generalisation under such non-stationary environment. However, current modular approaches in NLP do not take advantage of recent advances in parameter efficient tuning of pretrained language models. To close this gap, we propose \textsc{\small{ModularPrompt}}, a label-modular prompt tuning framework for text classification tasks. In  \textsc{\small{ModularPrompt}}, the input prompt consists of a sequence of soft \emph{label} prompts, each encoding modular knowledge related to the corresponding class label. In two of most formidable settings,  \textsc{\small{ModularPrompt}} outperforms relevant baselines by a large margin demonstrating strong generalisation ability. We also conduct comprehensive analysis to validate whether the learned prompts satisfy properties of a modular representation.\footnote{Our Code is available at \url{https://github.com/salesforce/ModularPrompt}}

\end{abstract}

\input{introduction.tex}

\input{related_work.tex}
\input{methods.tex}

\input{experiments.tex}

\section{Conclusion}

In this paper, we have proposed \mprompt, a novel label modular prompt tuning framework for text classification \textit{in-the-wild}. Extensive experiments show that \mprompt is able to consolidate knowledge learned during sequential training stages for stage-agnostic testing and extract and recompose knowledge for stage-fused testing, while maintaining competitive performance in stage-specific settings. We have also conduct analysis to show that \mprompt has desirable modular properties of label grounding, low order sensitivity and in-context learning. 
Being the first work on modular parameter efficient tuning, we hope for it to spur more research in this area in future towards solving a wider range of tasks under more general non-stationary settings.

\section*{Limitations}

In this section, we discuss limitations and potential future work towards extending \mprompt to a more generalised method for wider applicability.

\paragraph{On Scalability} In \mprompt, the input prompt $\T$ grows in proportion to $|S|$, the size of label set of interest. This limits \mprompt from supporting huge label set (e.g.,  thousands of labels) as transformers
can only condition on a bounded-length context. With long range transformers like Longformer \cite{Longformer}, Performer \cite{Performer} and LongT5 \cite{LongT5} coming into vogue, this issue is somewhat mitigated. Regardless of that, one potential solution is to formulate a hierarchical version of \mprompt, which is similar in spirit to hierarchical softmax \cite{hierarchical_softmax}. Hierarchical \mprompt takes multiple steps for prediction, with each step to predict labels in a specific hierarchy level. 

Another potential solution is to treat all label prompts as memory units from which the model learns to select relevant ones for a given data instance, in the spirit of \cite{memorizing_transformers}.

\paragraph{On generation tasks}
As \mprompt shows SoTA performance and good modular characteristics for text classification \textit{in-the-wild}, it is appealing to extend it to other tasks like Question Answering (QA), Machine Reading Comprehension (MRC) and Summarization. However, it is non-trivial for \mprompt to incorporate these tasks as their target texts are unstructured without clear class labels. One potential solution is to instead consider attributes or properties of target texts, which are also conditioning factors (e.g., formality, conciseness, topics, aspects, sentiment for summarization). With such definitions, it will be interesting to check if \mprompt framework can achieve good generalisation and conditional generation on text generation in-the-wild.

\bibliography{emnlp2022}
\bibliographystyle{acl_natbib}

\newpage
\input{appendix}

\end{document}

%% file: introduction.tex
\section{Introduction} \label{sec:intro}


While NLP research has received a significant boost in performance by employing large-scale pretrained language models (PLMs), finetuning an entire dedicated model for each task is not always practical  or even feasible, especially as model size continues to grow. To alleviate this, recently there has been increased interest in parameter-efficient methods such as {Adapter} \cite{adapter} and Prompt-based tuning methods \cite{soft-prompt,Mix_soft_prompts,p_tuning,prefix_tuning}. When training on a downstream task, these methods keep the PLM frozen  and only update a small set of parameters that are added to the network. Among these methods, \textsc{\small{PromptTuning}} \cite{soft-prompt} proposes \emph{tunable prompts} -- a sequence of learnable soft tokens, and show impressive results, even competitive  to full-model finetuning with a large PLM.



While these prompt-based methods have  proven quite effective on popular benchmarks, these standard tasks typically assume only independently and identically distributed (i.i.d) data during training and testing. However, practical cognitive tasks in the real world are usually more complex involving changing contexts or non-stationary environments. Talking about what is next for NLP, Kathleen McKeown in a recent interview  (\href{https://www.amazon.science/blog/acl-what-comes-next-for-natural-language-processing}{source}) said: 
``Most models are static. But the world changes every minute, every second. Dealing with a dynamic world is a new area that's up and coming.''  


\begin{figure}
	\centering
	\includegraphics[width=0.48\textwidth]{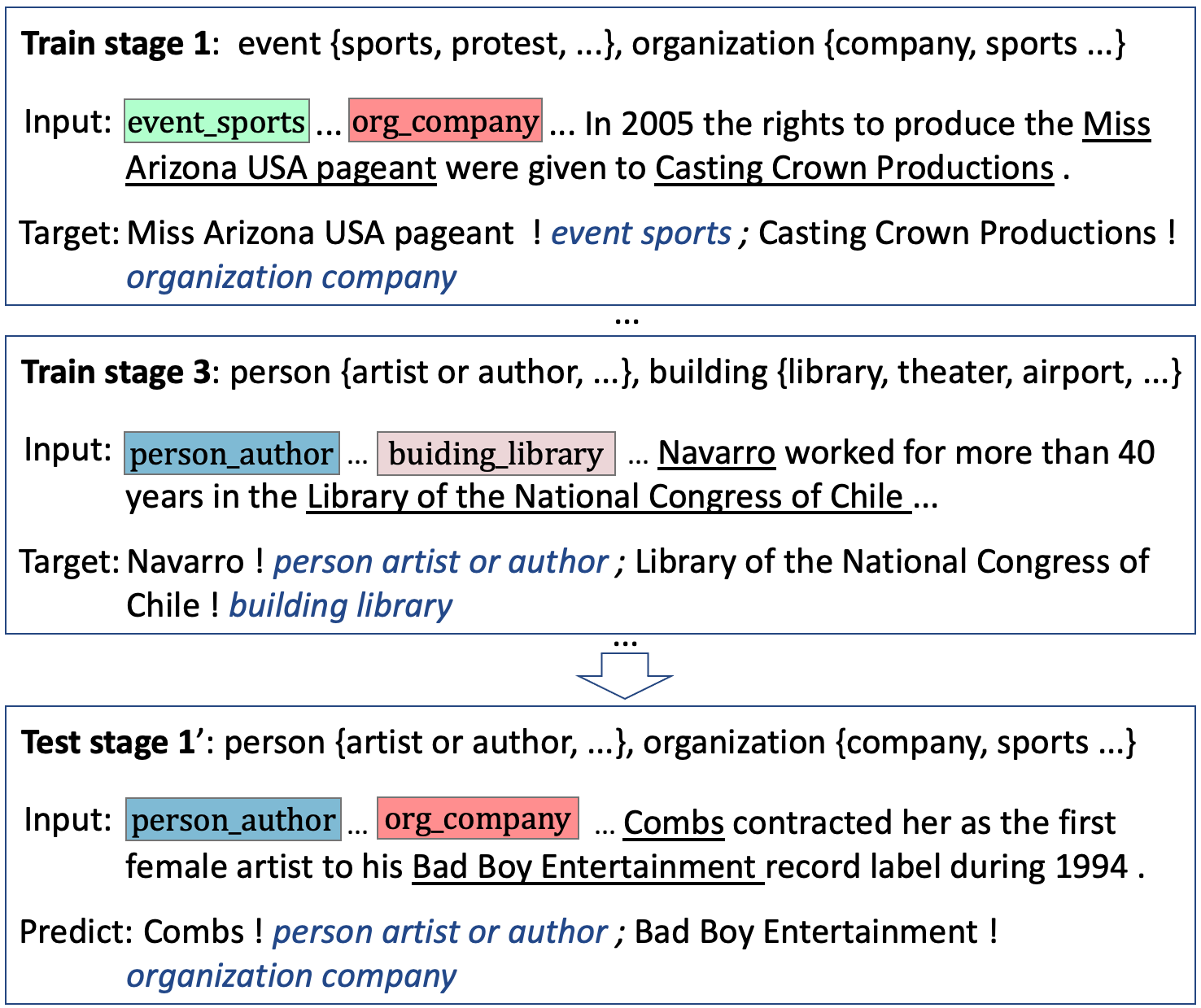}
	\caption{\small Example of \mprompt for stage-fused NER. \textbf{Top 2 blocks}: training stages covering disjoint sets of entity types {(e.g., \emph{event, person})}. \textbf{Bottom block}: fused test stage covering entity types of \emph{person} and \emph{organization} from two training stages. \colorbox{lightPink}{Coloured boxes} denote label prompts. \underline{Underlining} and \textit{\darkBlue{italic blue}} denotes named entity and its type.} 
	\label{fig:case_study}
\end{figure}

Our work in this paper particularly concerns text classification settings where a model is trained on sequence of tasks and evaluated on an arbitrary subset of seen labels of interest.
We formalize this as a novel \textbf{text classification \textit{in-the-wild}} task (defined in \Cref{sec:method}), which emulates the transient learning environment of real world, e.g., for a service requiring classification, the label set might gradually change over time to include new labels or remove obsolete ones. Such scenarios typically result in a sequence of non-stationary low-resource training and evaluations over different label sets  (e.g., train on \{\emph{chemistry, physics}\} and \{\emph{basketball, football}\} in succession and then test on \{\emph{physics, football}\}).


This requires handling non-stationary data distribution which humans are quite adept at, partly because we can decompose a complex task in a modular fashion \cite{Berwick}. For example, when learning to classify objects, we acquire modular knowledge exclusive to each class. This allows us to robustly classify irrespective of any label space manipulations such as label omission or learning over new label spaces. This notion of modularity at the level of each class label is what we call \emph{label modularity} and is a desirable quality for NLP models to generalize to practical non-stationary classification settings.


Contemporary modular model designs for complex NLP tasks typically use a routing network or programmer \cite{RecursiveRouting,RoutingNetworks,TextModular,ModularInstruction,ModularMultiHop,ModularNMNtree,StackNMN,TextNMN,NMN} which learns a meaningful decomposition of the task into sub-tasks and executes it by applying a chain of specialised modules designed for each sub-task. 
For classification tasks, this can entail learning a specialized module for each label in the target label-space. While there has been limited research on utilizing modular architectures for PLMs \cite{NMN, NeRD}, the main research gap that we explore in this work is that of a modular design of parameter efficient tuning of large PLMs, in particular, \textsc{\small{PromptTuning}}. 

Although \textsc{\small{PromptTuning}} can be considered modular at task level in that it learns soft-prompts for each task to support multitasking, it is not able to learn modular decomposition within a particular task. For non-modular designs like \textsc{\small{PromptTuning}} or model-finetuning, text classification \textit{in-the-wild} is challenging to handle, as it requires combining partial information from different label spaces. In contrast, a label-modular approach should learn exclusive knowledge for each label and generalise to any subset of the label set. We thus postulate two main objectives of a label-modular model:

\begin{objective} \label{obj:sl}
	\textbf{Separable Label Representation}: Each class label should have its own representation which compactly encodes the information from the data belonging to that label.
\end{objective}
\begin{objective} \label{obj:cl}
	\textbf{Prediction over Controllable Label Space}: Models should perform robustly over any subset of the learnt label space during inference.
\end{objective}

\noindent To meet these objectives, we propose a modular design of Prompt Tuning -- \textbf{Label-modular Prompt Tuning} (\mprompt). It decomposes the prompt sequence into label-modular components called \emph{label prompts}, each encoding specific knowledge corresponding to a class label. Thus in each forward pass, we can select desired label prompts  to construct the input prompt, based on the target label-set. To ensure that the learned knowledge is encoded in a modular fashion during training, we introduce a novel subset-invariant loss over dynamic label-sets. 



To evaluate generalizability of \mprompt\, we construct some practical scenarios of text classification \textit{in-the-wild}. 
We train in multiple stages over non-overlapping label spaces and evaluate the model on label-sets that (i) correspond to each training stage (stage-specific),  
(ii) is accumulated over all learned labels (stage-agnostic), and (iii) is comprised of labels across multiple training stages (stage-fused). 
We show an example of \mprompt on stage-fused NER setting in  \Cref{fig:case_study}. 

The stage-agnostic and stage-fused settings are the most challenging scenarios for typical fine-tuned or prompt-tuned models, and specifically on those settings we find that \mprompt\ outperforms all relevant baselines 
by a significant margin. This alludes towards its ability to learn robust prompt representations that is generalizable to different non-stationary learning environments.

We further empirically justify that \mprompt\ indeed showcases modular properties by analyzing its behavior when either the ground truth or other random labels are removed from the input or the order of label prompts is permuted.

%% file: related_work.tex
\section{Related Work}

We now review methods from the literature that are relevant to ours from different perspectives.
First, many parameter efficient tuning methods have been proposed such as Adapter \cite{adapter}, \textsc{\small{PromptTuning}} \cite{soft-prompt}, \textsc{\small{PrefixTuning}} \cite{prefix_tuning}, BitFit \cite{bitfit}, LoRA \cite{lora} and COMPACTER \cite{compacter}. They either finetune a parameter subset or introduce new layers or embeddings. 
Transfer learning over prompts \cite{SPoT} and Adapters \cite{adapterFusion} and multi-tasking over the latter \cite{hyperadapter} have also been explored, but they are not comparable to ours due to difference in task/problem settings.  

Second, continual learning (CL) methods are relevant as they also have sequential training stages. Architecture based CL methods adjust model architecture for each task \cite{Net2Net,progressiveNN,piggyback}, but require task identities for inference. Regularization based CL methods restrain updating parameters critical to previous tasks \cite{EWC,synapticCL,MAS}. Memory based CL methods retain key examples from prior tasks \cite{GEM, AGEM, episodic_memoty, continualPT} while Memory generator models learn to generate and use pseudo-data from prior tasks \cite{LAMOL,lfpt5}. 
These methods are not comparable to ours as we do not use any memory or pseudo memory in our sequential training.

Third, modular networks have been shown to perform well on out-of-domain data \cite{ModularNetworks,LatentMTAL,ModularMeta} and mitigate forgetting in continual learning \cite{ModularCL}.  Apart from routing network approaches introduced in \Cref{sec:intro}, mixture-of-experts (MoE) selects a soft subset of modules based on model input \cite{SparseMoE,SwitchTransformers,RIM}. Similar to ours, \citet{{TaskMoE,AttentiveTransfer,FactorZeroShot,ModularCL}} consider task level routing and support new tasks by combining learned modules. By contrast, \mprompt utilises parameter efficient finetuning of large PLMs and support low resource settings.

%% file: methods.tex
\section{Methodology} \label{sec:method}
In this section, we first formally define the problem and subsequently present our \mprompt model, introduce subset invariant loss and explain the framework under text classification \textit{in-the-wild}. 

\begin{figure*}[htbp]
	\centering
    \vspace{-5pt}
    \small
	\setlength{\abovecaptionskip}{2pt} 
	\setlength{\belowcaptionskip}{-10pt} 
	\includegraphics[width=1.0\textwidth]{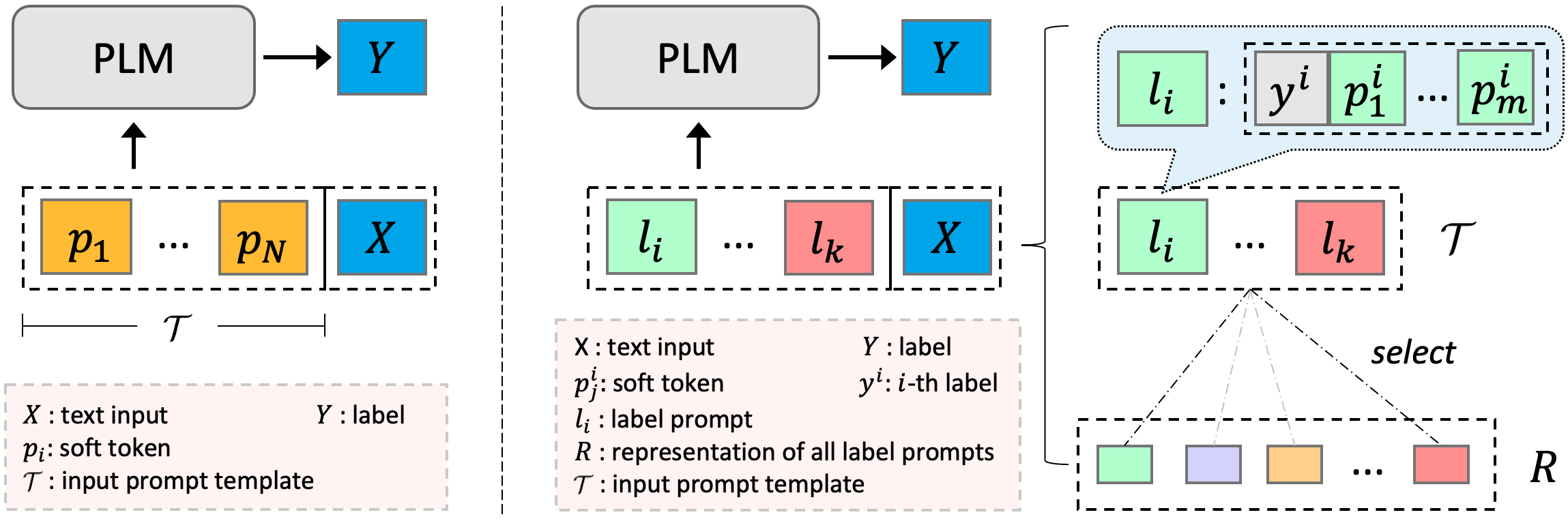}
	\caption{\textbf{Left}: \textsc{\small{PromptTuning}}, where consecutive soft tokens are concatenated with the input to a frozen PLM. \textbf{Right}: Our proposed \mprompt, where multiple label prompts are concatenated together with the input to a frozen PLM. Each label prompt consists of a label name and consecutive sequence of soft tokens.} 
	\label{fig:model}
\end{figure*}

\subsection{Problem Definition}

\paragraph{Single stage text classification} \label{single-domain}
Assume a single text classification domain (or dataset) $\D$. Let $(X, Y) \sim \D$ be a sample, where $X=\{x_t\}_{t=1}^L$ represents a text input sequence of length $L$ and $Y=\{y_t\}_{t=1}^M$ represents the corresponding classification label name of length $M$ (in tokens). Let $\Omega$ denote the set of all possible class labels of interest, for which we have $\forall (X,Y) \sim \D, \text{cls}(Y) \subseteq \Omega$. Note that $\text{cls}(Y)$ is a mapping which returns the class label(s) in $Y$. In case of single class classification, $\text{cls}(Y)$ returns $\{Y\}$. In case of sequence labelling which is token-level classification, $\text{cls}(Y)$ returns the set of all unique target tags in $Y$.

\vspace{-0.5em}
\paragraph{Text classification in-the-wild} \label{in-the-wild}

Assume a sequence of $n$ text classification stages with the corresponding training datasets $\DD^{tr}=\{\D^{tr}_1, ...,\D^{tr}_n\}$.
Each stage represents a different task in temporal dimension, with $(X_k, Y_k) \sim \D^{tr}_k$ denoting a sample at the k-th training stage and $\Omega_k$ denoting the set of all possible class labels for $\D^{tr}_k$. Similarly, the testing could consist of $m$ such datasets $\DD^{ts}=(\D^{ts}_1,  ...,\D^{ts}_m)$ with $\Omega^{ts}_{j}$ denoting  the set of possible class labels for $\D^{ts}_j$. For classification \textit{in-the-wild}, we examine three challenging yet very practical settings. First, when $m = 1$ and $\Omega^{ts}_{1} = \cup_{k=1}^n\{\Omega^{tr}_k\}$, we have one test dataset covering all seen labels. We refer it as \textbf{stage-agnostic} testing as the test label can come from any of the training stages (or tasks). Second, when $m = n$ and $\Omega^{ts}_{j} = \Omega^{tr}_j, \forall j =\{1,...,n\}$, we have one test dataset corresponding to each training stage with the same label set. We denote this setting as \textbf{stage-specific} testing as each test set evaluates the model's performance on a particular task in which it was trained. Finally, a more challenging setting where $m >1$ and $\Omega^{ts}_{j}\notin \{\Omega^{tr}_1,...,\Omega^{tr}_n\}, \forall j =\{1,...,m\}$, rather $\Omega_j^{ts} \in \mathcal{P}(\cup_{k=1}^n\{\Omega^{tr}_k\}) - \cup_{i=1}^n\{\mathcal{P}(\Omega^{tr}_i)\}$, where $\mathcal{P}(S)$ denotes the power-set of a given set $S$. That is, the label set of a test stage does not correspond to any one training stage, but is composed of partial label sets from multiple training stages (or tasks). We refer it as \textbf{stage-fused} testing. Note that the stage-agnostic and stage-specific scenarios are closely related to \emph{continual learning} \cite{Thrun96}, though the latter considers access to task-id instead of intra-task information (i.e., task label set).

\subsection{Soft Prompt Tuning} \label{soft-prompt}

Let $X = \{x_1, ..., x_L\}$ be an input text sequence, where $x_t$ is the $t$-{th} token, and $M$ be a pretrained language model. The input text is mapped to a sequence of embeddings $H = \{\bm{h}_1, ..., \bm{h}_L\}$ with $\bm{h}_t \in \R^d$. A soft prompt is a sequence of $N$ tunable tokens $\T = \{\bm{p}_1, ..., \bm{p}_N\}$ with $p_i \in \R^d$, that is concatenated with the text embedding as the final input to $M$: $\bar{H} = \{\T \oplus H\}=\{\bm{p}_1, ..., \bm{p}_N, \bm{h}_1, ..., \bm{h}_L\}$. The model prediction is defined as $P(Y|\bar{H};M)=P(Y|\T,X;M)$. During training, $M$ is kept frozen and only $\T$ is updated.

\subsection{Label Modular Prompt Model} \label{sec:mprompt}

We visualise the architecture of \textsc{\small{PromptTuning}} \cite{soft-prompt} and our proposed \mprompt\ in \Cref{fig:model}. In contrast to \textsc{\small{PromptTuning}}, \mprompt's prompt consists of a sequence of label prompts, where each label prompt contains the corresponding label name and a sequence of tunable soft tokens similar to soft prompt. Formally, we denote $\bm{l}_{k} = \bm{e_k} \oplus \{\bm{p}^k_1, ..., \bm{p}^k_m\}$ as label prompt for label $k$, where $\bm{e_k}$ is the embedding of label $k$'s text or sequence of token-embeddings for multi-token labels, $\oplus$ denotes concatenation and $m$ is the number of tunable tokens per label prompt. The final input prompt is $\T = \oplus_{k \in S} \bm{l}_{k}$ with $S$ being the set of labels of interest.



\vspace{-0.5em}
\paragraph{Prompt formulation} 
The key mechanism of \mprompt is prompt formulation $\{R, S\} \rightarrow \T$, where $R$ denotes the learned representation space of all labels prompts. In \textsc{\small{PromptTuning}}, variables $S$ and $R$ do not exist and the model training tunes $\T$ directly. In \mprompt, given $S$ as a set of class labels of interest, we select the corresponding label prompts representation from $R$ and concatenate these to form the final input prompt $\T$. 
The training loss is back-propagated through $Y \rightarrow \T \rightarrow R$ to learn the soft label prompts.

\paragraph{Subset invariant loss}

The prompt formulation $\{R, S\} \rightarrow \T$ aims to achieve Objective \ref{obj:cl}: prediction over controllable label space (\Cref{sec:intro}). In single domain setting, $\SSS^{tr}$ is the set of all possible class labels during training as defined by Section \ref{single-domain}.
However fixing $S$ to a constant $\SSS^{tr}$ throughout training will make the model susceptible to data discrepancy between train and inference as $\SSS^{ts} \neq \SSS^{tr}$. Thus to ensure Objective \ref{obj:cl}, we propose to vary $S$ during training. We first uniformly sample the size of $S$, $|S|$ from $\{1, \ldots, (|\Omega^{tr}|-1) \}$ and then randomly choose $|S|$ labels from $\Omega^{tr}$ to construct $S$. Such sub-sampling of {$\Omega^{tr}$} encourages a fair exploration of different lengths of prompt sequences as input during training, thus enabling representations to be robust to a dynamic $\Omega^{ts}$ at inference. For each training instance, with probability $p$ we fix $S=\Omega^{tr}$ and vary $S$ as above with $(1-p)$ chance. 
We refer such sampling process as $S \sim \hat{\mS}$. 
The subset invariant loss is then defined as:
\begin{equation}\small
	\LL^{\text{inv}}_{R} (\D) = \mathbb{E}_{\substack{(X,Y) \sim \D \\ S \sim \hat{\mS}}} \big[ - \mathbbm{1}_{\text{cls}(Y) \subseteq S} \log  P(Y|X, R, S; M) \big] 
\label{eq:subset_inv}
\end{equation}
where $\mathbbm{1}$ is the Indicator function;
$\mathbbm{1}_{\text{cls}(Y) \subseteq S} = 1$ if $\text{cls}(Y) \subseteq S$, otherwise $0$. According to Objective \ref{obj:sl}, we expect our model to make predictions grounded by the relevant label prompts. When $S$ does not contain  ground truth class label(s) in $Y$, the model should not be able to predict $Y$ as output. Thus we set the loss to be zero when $\text{cls}(Y) \nsubseteq S$ to avoid encouraging ungrounded predictions.

\subsection{\hspace{-0.5em} Classification \textit{in-the-wild} with \mprompt } \label{sec:mprompt_in_the_wild}
So far, we have introduced \mprompt\'s functionality under a single domain classification setting. To verify our Objective \ref{obj:cl}, we wish to examine it under text classification \textit{in-the-wild} defined in \Cref{in-the-wild}. Given training datasets $\DD^{tr} = \{\D^{tr}_1, ..., \D^{tr}_n\}$, the model is trained on each dataset $\D^{tr}_i$ sequentially, and then evaluated on three classification \textit{in-the-wild} testing settings. The pseudocode of \mprompt\ under the continual learning setting is given in \Cref{alg:cl}. Note that $R_{\SSS^{tr}_{i}}$ in step 3 denotes label prompt representation of labels in $\SSS^{tr}_{i}$, i.e., $R_{\SSS^{tr}_{i}} \coloneqq \{\bm{l}_k \in R | k \in \SSS^{tr}_{i}\}$
and $R_{\SSS^{tr}_{<i}}$ is similarly defined as $R_{\SSS^{tr}_{<i}} \coloneqq \{\bm{l}_m \in R | m \in \underset{t<i}{\cup} \SSS^{tr}_t\}$. 

\begin{algorithm}[t]
\caption{Text Classification \textit{in-the-wild} with \mprompt.}\label{alg:cl}
\small 
\begin{algorithmic}[1]
\Require Training datasets $\DD^{tr} = \{\D^{tr}_1, ..., \D^{tr}_n\}$, testing datasets $\DD^{ts}=(\D^{ts}_1, ...,\D^{ts}_m)$, pretrained language model $M$
\For{ $\D^{tr}_i$ in $\DD^{tr}$} \Comment{Training}
	\State $\SSS^{tr} \leftarrow \SSS^{tr}_{i}$ 
	\State INITIALIZE $R_{\SSS^{tr}_{i}}$ from
	$R_{\SSS^{tr}_{<i}}$ \Comment{Transfer}
	\For{$step = 1,..., iter$}
		\State UPDATE $R_{\SSS^{tr}_{i}}$ with SGD on $\LL^{\text{inv}}_R (\D^{tr}_i)$
	\EndFor
\EndFor

\For{ $\D^{ts}_i$ in $\DD^{ts}$} \Comment{Testing} 
	\State EVALUATE $P(\cdot|X, S=\SSS^{ts}_{i},R; M)$ on $X \sim \D^{ts}_i$
\EndFor
\end{algorithmic}
\end{algorithm}

\paragraph{Label prompt transfer} 
In step 3, for learning the label prompt representation $R_{\SSS^{tr}_i}$ at any training stage $i$, we first aim to transfer the label-modular knowledge, $R_{\SSS^{tr}_{<i}}$ learned over the previous training stages through prompt initialization. This is a unique learning characteristic that is facilitated by our label-modular architecture and allows the model to exploit semantic relatedness between labels across training stages when initializing the label prompt representation. Intuitively, if `bistro' $\in \SSS^{tr}_{<i}$ and `restaurant' $\in \SSS^{tr}_i$, then initializing the label prompt representation of `restaurant' with the knowledge encoded in the learned label prompt representation of `bistro' should be  helpful to the model. To compute the similarity between labels $\bm{l}_j$ and $\bm{l}_k$ with $j \in \SSS^{tr}_i$ and $k \in \SSS^{tr}_{<i}$, we use per-token average cosine similarity $sim(\bm{e}_{j}, \bm{e}_{k})$ based on the embeddings of the label texts. For each label $j \in \SSS^{tr}_{i}$, we select the top-$K$ most similar labels $\SSS^{tr}_{\text{top-}K{(j)}} \subset \SSS^{tr}_{<i}$. We then initialize $\bm{l}_j$ by averaging the top-$K$ similar label prompt representations, weighted by their normalized similarity score: $\bm{l}_j \leftarrow \sum_{k \in \SSS^{tr}_{\text{top-}K{(j)}}} \alpha_{k} \bm{l}_k $, where $\alpha_{k} = {sim(\bm{e}_{j}, \bm{e}_{k})} / {\sum_{m \in \SSS^{tr}_{\text{top-}K{(j)}}} sim(\bm{e}_{j}, \bm{e}_{m})}$. This  method is similar in spirit to \cite{SPoT}, which shows good transfer for task level prompts with training overheads, while we transfer at a finer-grained level over label prompts with no overheads.




%% file: experiments.tex
\section{Experiments}
In this section, we first introduce datasets used and data construction process (\Cref{sec:exp:tasks}) followed by relevant baselines (\Cref{sec:exp:baselines}), evaluation methods (\Cref{sec:exp:eval}) and implementation details (\Cref{sec:exp:imp}). Through our experiments, we target three research questions:
\begin{enumerate}[leftmargin=*,topsep=2pt,itemsep=2pt,parsep=0pt]
    \item Can \mprompt consolidate knowledge over multi-stage training? $\rightarrow$ answered in \Cref{sec:exp:agnostic} with stage-agnostic setting
    \item Can \mprompt adapt to dynamic label space at inference? $\rightarrow$ answered in \Cref{sec:exp:fuse} with stage-fused setting
    \item How competitive is \mprompt in stage-specific setting? $\rightarrow$ answered in \Cref{sec:exp:specific}
\end{enumerate}
Additionally, we perform ablations (\Cref{sec:exp:ablation}) and  quantitative and qualitative analysis (\Cref{sec:exp:quant_analysis}-\Cref{sec:exp:qual_analysis}) to verify modular properties of \mprompt.

\subsection{Tasks and Datasets} \label{sec:exp:tasks}
We conduct experiments on three types of NLP tasks: News  Domain Classification on HuffpostNews \cite{huffpost, misra2021sculpting}, Name Entity Recognition (NER) on fewNERD \cite{fewNERD}  and Relation Extraction (RE) on FewRel \cite{fewrel}. We formulate all tasks as a text-to-text problem, as defined in \Cref{in-the-wild}. 
For News Domain Classification and NER, we construct target text following \citet{lfpt5}. 
For RE, we concatenate the original text and entities with a seperator '|' as the input sequence, and use the relation type as the target. (Example data can be found at \cref{appendix:dataset})

For HuffpostNews, we subsample 100 shots per class for training and validation and split it into 5 stages of disjoint labels. For FewNERD and FewRel, we subsample 50 shots for training and validation and split into 4 and 5 stages, respectively. For testing, we subsample 200, 50, and 50 shots per class for HuffpostNews, FewNERD and FewRel, respectively. The total number of labels for \{HuffpostNews,FewNERD,FewRel\} is \{41,64,80\} respectively, and resulting label size per stage is \{8-9,16,16\} respectively. 

For stage-specific testing, we follow the stages defined for training and construct a corresponding test data for each stage. For stage-agnostic testing, we combine stage-specific test data for current stage and all previously seen stages to construct the test data. For stage-fused testing, we construct label-sets for each fused stage such that it is not a subset of any single \text{prior} training stage, but rather contains labels from `all' {prior} training stages. 
We construct \{5,4,5\} fused stages for \{HuffpostNews,FewNERD,FewRel\}. We conduct 5 randomised trials with different data sampling and experiment seed for all of the above settings. 

\subsection{Baselines} \label{sec:exp:baselines}
We use T5-large \cite{T5} as the backbone PLM for all methods, and consider the following baselines to compare with our \textsc{\small{ModularPrompt}}: 
\noindent
\begin{itemize}[leftmargin=*]
\vspace{-0.5em}
	\item \textsc{\small{ModelTuning}} (Finetune), which tunes all parameters of the backbone PLM.
\vspace{-0.5em}
	\item (i) \textsc{\small{PromptTuning}} (PT) from  \Cref{soft-prompt},  (ii) PT$_{\text{CL}}$ - An extension of PT to continual learning (CL) setting, which trains separate PT models for each stage and concatenates the learned soft-prompts during inference, based on the test label-set. 
\vspace{-0.5em}
	\item Adapter, a parameter efficient tuning alternative introduced in \cite{adapter}, which inserts light adapter layers into the backbone PLM and only tune them. 
\end{itemize}
\vspace{-0.5em}
As text classification \textit{in-the-wild} overlaps with continual learning, we also compare with versions of the above baselines that use architecture-agnostic methods and settings relevant to the latter. 
\noindent
\begin{itemize}[leftmargin=*]
\vspace{-0.5em}
    \item Online regularization based methods: (i) A scalable online version of EWC \cite{EWC} proposed in \cite{onlineEWC}, and (ii) Online MAS \cite{MAS}. These methods measure the importance of parameters to previous tasks by fisher information, and restrict updating previously important parameters when learning a new task, to mitigate catastrophic forgetting.
\vspace{-0.5em}
    \item Multitask model, which involves training on all stages simultaneously, not sequentially. This is infact an oracle method for stage-agnostic testing and can be considered as an upper bound of memory-based methods in continual learning. 
\end{itemize}

\subsection{Evaluation Methods} \label{sec:exp:eval}
For all the three NLP tasks, we consider an \emph{exact match} as a correct prediction and report accuracy for News Classification and RE, and compute F1-score over the BIO format for the NER task. By default, we do not apply any other post-processing or verbalizer, though these are orthogonal methods that can be separately used to enhance any of the discussed models. In the stage-fused setting, we apply constrained decoding similar to \cite{constrained_decode} to selected baselines, marked by special indicator * (e.g., Finetune$^*_{\text{MAS}}$). For \textsc{\small{ModularPrompt}}, we use all seen label prompts for stage-agnostic testing and specific set of label prompts for stage-specific and stage-fused testing. Since other baselines do not have label-level modularity,  for stage-agnostic and stage-fused testing, we use the checkpoint after the final stage and for stage-specific testing we take their checkpoints after each training stage. We show average performance in the main paper and relegate detailed results to Appendix \ref{sec:appendix_results}.

\subsection{Implementation Details} \label{sec:exp:imp}
We set the learning rate to 0.5 for \textsc{\small{PromptTuning}} and \textsc{\small{ModularPrompt}} and 5e-5 for \textsc{\small{ModelTuning}} and Adapter, using Adafactor \cite{adafactor} optimizer. We adopt implementation of Adapter from OpenDelta \cite{OpenDelta} and use the default bottleneck dimension of 24. For online EWC and MAS, we report best results obtained over different regularization constant. For all methods, we set maximum training epochs to 256 for FuffpostNews and FewNERD, and to 512 for FewRel. For \textsc{\small{ModularPrompt}}, the number of soft tokens per label prompt is set to 10, the selection probability $p$ is set to 50\% and number of label transfer candidates K in \Cref{sec:mprompt_in_the_wild} is set to 3.

\subsection{Results on Stage-agnostic Setting} \label{sec:exp:agnostic}
In Table \ref{table:agnostic:all}, we show the stage-agnostic testing results. 
We observe that across all three tasks,  \textsc{\small{ModularPrompt}} significantly outperforms all other baselines by a large margin. This empirically justifies that \textsc{\small{ModularPrompt}} is indeed able to dynamically combine the label-specific knowledge learned across different training stages in order to infer over the \emph{unseen} combined label-space. Amongst the baselines, \textsc{\small{ModelTuning}} performs relatively better, while the limited trainable parameters make the parameter efficient models more susceptible to catastrophic forgetting. 
{For CL methods, MAS improves \textsc{\small{ModelTuning}} and \textsc{\small{PromptTuning}} by 4\% and 8\% on average respectively, but fails on Adapter. EWC is less effective in addressing forgetting across all baselines.}


Also note that the PT$_{\text{CL}}$ extension is able to improve by 10-20\% over vanila PT. This shows that soft prompts, behaving like language tokens, have a compositional nature and can be concatenated to support multi-tasking. \textsc{\small{ModularPrompt}}, in addition to exploiting this implicit language prior, also explicitly imposes subset-invariant loss to adapt to dynamic label spaces, further boosting stage-agnostic performance by 14\%-18\% over PT$_{\text{CL}}$.

\begin{table}
\centering
\scalebox{0.75}{
\begin{tabular}{llccc}
\toprule
\textbf{Methods} & \textbf{Size}   & News & NER & RE\\ 
\toprule
Finetune$_{\text{multitask}}$     & 770M &    60.6$_{\pm 5.0 }$    &    64.5$_{\pm 0.2 }$    &    87.9$_{\pm 0.6 }$\\[0.5ex]
\midrule
Finetune      & 770M &    23.0$_{\pm 0.6 }$    &    25.0$_{\pm 2.7 }$    &    40.8$_{\pm 2.2 }$\\ 

Finetune$_{\text{EWC}}$    & 770M &    22.5$_{\pm 1.2 }$    &    26.6$_{\pm 2.8 }$    &    45.4$_{\pm 3.2 }$ \\

Finetune$_{\text{MAS}}$    & 770M &   29.2$_{\pm 2.9}$    &    25.4$_{\pm 2.9 }$    &    46.2$_{\pm 1.8 }$\\

Adapter    & 4.8M &    19.3$_{\pm 0.3 }$    &    21.2$_{\pm 2.0 }$    &    28.6$_{\pm 1.9 }$\\

Adapter$_{\text{EWC}}$    & 4.8M &   18.6$_{\pm 1.0}$    &   20.5$_{\pm 1.3}$    &   27.8$_{\pm 1.2}$  \\

Adapter$_{\text{MAS}}$    & 4.8M &    20.1$_{\pm 1.5}$    &    21.1$_{\pm 2.0}$    &    28.3$_{\pm 0.5}$ \\

PT    & 0.13M &    16.1$_{\pm 0.3}$    &    18.8$_{\pm 0.1 }$    &    19.6$_{\pm 0.1 }$ \\

PT$_{\text{EWC}}$    & 0.13M &    16.5$_{\pm 0.5}$     &    19.0$_{\pm 0.5}$     &   20.1$_{\pm 0.8}$ \\

PT$_{\text{MAS}}$    & 0.13M &    23.8$_{\pm 5.9}$    &    29.3$_{\pm 4.6}$    &    25.1$_{\pm 4.9}$  \\

PT$_{\text{CL}}$    & 0.13M &    28.1$_{\pm 2.9 }$    &    30.9$_{\pm 5.4 }$    &    43.1$_{\pm 4.9 }$\\[0.5ex]
\hline
ModularPT          & 0.13M &    \textbf{43.2}$_{\pm 0.6 }$    &    \textbf{44.8}$_{\pm 4.9 }$    &    \textbf{61.8}$_{\pm 1.8 }$\\
\bottomrule
\end{tabular}
}
\caption{\textbf{Stage-agnostic} performance on News Classification, NER and Relation Extraction (RE). \textbf{Size} denotes average number of tunable parameters per training stage}
\label{table:agnostic:all}
\end{table}

\subsection{Results on Stage-fused Setting} \label{sec:exp:fuse}
We present results on our novel stage-fused setting in Table \ref{table:fuse:all}. We observe that none of the baselines are capable of handling this setting, as is evident from their abysmal performance across all testing stages. In absence of any label-modular representation, they are unable to utilize any information about the desired label-space. On the other hand, \textsc{\small{ModularPrompt}} not only outperforms all baselines by an average margin of 37.5\%, it also achieves 4\%-14\% better performance than the oracle multi-task \textsc{\small{ModelTuning}} on News Classification and NER. 

We select the top performing baselines  in this setting and apply constrained decoding to them (marked with *), which improves their performance by 20\%-30\% on News and RE, 2\%-4\% on NER. However, \mprompt still outperforms these baselines by 14\%-27\%. This significant improvement is evident of the fact that \textsc{\small{ModularPrompt}}, by learning label-modular representations, can effectively combine partial knowledge from different training stages and condition the PLM on any target set of label prompts. This allows it to seamlessly adapt to dynamic unseen label spaces, without applying any post-processing or verbalizer.

\begin{table}[t!]
\centering
\scalebox{0.75}{
\begin{tabular}{llccc}
\toprule
\textbf{Methods} & \textbf{Size}     & News & NER & RE\\ 
\toprule  
Finetune$_{\text{multitask}}$     & 770M &    60.7$_{\pm 0.6 }$    &    57.0$_{\pm 0.0 }$   &   86.0$_{\pm 0.9 }$\\[0.5ex]
\hline
Finetune      & 770M &    22.4$_{\pm 0.5 }$    &    24.9$_{\pm 2.1 }$    &   40.7$_{\pm 2.1 }$\\ 

Finetune$_{\text{EWC}}$    & 770M &    22.6$_{\pm 1.0 }$    &    25.1$_{\pm 1.7 }$   &   45.2$_{\pm 3.3 }$\\

Finetune$_{\text{MAS}}$    & 770M &    28.8$_{\pm 3.3 }$    &    23.5$_{\pm 1.3 }$   &   46.9$_{\pm 1.9 }$\\

Adapter    & 4.8M &    18.8$_{\pm 0.3 }$     &    21.8$_{\pm 1.7 }$   &   28.2$_{\pm 1.8 }$\\

Adapter$_{\text{EWC}}$    & 4.8M &   18.1$_{\pm 1.0}$     &   21.7$_{\pm 1.5}$   &  27.4$_{\pm 1.2}$  \\

Adapter$_{\text{MAS}}$    & 4.8M &   19.6$_{\pm 1.4}$     &   21.4$_{\pm 2.0}$    &  27.8$_{\pm 0.6}$ \\

PT    & 0.13M &    15.7$_{\pm 0.2 }$     &    19.3$_{\pm 0.5 }$   &   19.6$_{\pm 0.2 }$ \\

PT$_{\text{EWC}}$   & 0.13M &   16.1$_{\pm 0.5}$   &   20.0$_{\pm 1.0}$  &  20.0$_{\pm 0.8}$ \\

PT$_{\text{MAS}}$    & 0.13M &   23.7$_{\pm 6.2}$     &   25.3$_{\pm 3.8}$   &  24.9$_{\pm 5.1}$ \\

PT$_{\text{CL}}$    & 0.13M &    27.5$_{\pm 2.8 }$     &    30.4$_{\pm 5.1 }$   &   43.3$_{\pm 5.2 }$\\[0.5ex]
\hline
Finetune$^*_{\text{MAS}}$ & 770M  &  52.8$_{\pm 2.2}$   &  25.8$_{\pm 1.7}$ & 68.1$_{\pm 1.3}$\\

PT$^*_{\text{CL}}$ & 0.13M  &  57.5$_{\pm 5.2}$   &  34.3$_{\pm 2.4}$ & 65.3$_{\pm 2.7}$\\[0.5ex]
\hline
ModularPT          & 0.13M &    \textbf{74.8}$_{\pm 1.7 }$    &    \textbf{61.6}$_{\pm 2.2 }$   &   \textbf{82.4}$_{\pm 0.8 }$ \\
\bottomrule
\end{tabular}
}
\caption{\textbf{Stage-fused} performance on News Classification, NER and relation extraction (RE)}
\label{table:fuse:all}
\end{table}

Note that while PT$_{\text{CL}}$ is able to combine knowledge from multiple training stages to support stage-agnostic testing, it fails to extract and consolidate specific knowledge corresponding to only the target label-set, across different stages.

\subsection{Results on Stage-specific Setting} \label{sec:exp:specific}

While \textsc{\small{ModularPrompt}} has proved to be particularly successful in handling the challenging non-stationary settings of stage-agnostic and stage-fused evaluations, we now want to see how competitive it is under stage-specific settings. From the results in Table \ref{table:specific:all}, we see that the average stage-specific performance of \textsc{\small{ModularPrompt}} is comparable to vanila \textsc{\small{PromptTuning}} on the three tasks. 
Note that while MAS regularization boosts stage-agnostic performance somewhat for \textsc{\small{ModelTuning}} and \textsc{\small{PromptTuning}}, it infact degrades their stage-specific performance by 10\%-40\%. Similarly applying EWC regularization fails to improve over the vanila models in this setting while also proving less effective on stage-agnostic evaluation. This shows the lack of robustness of these techniques across the different non-stationary settings. But \textsc{\small{ModularPrompt}} is able to achieve state-of-the-art in stage-agnostic and stage-fused settings while remaining comparable to \textsc{\small{PromptTuning}} in stage-specific evaluation. 
{Besides, \cite{soft-prompt} showed that the performance gap between \textsc{\small{PromptTuning}} and \textsc{\small{ModelTuning}} will gradually close as the size of backbone PLMs scales up.}
We posit that \textsc{\small{ModularPrompt}}, being an extension of \textsc{\small{PromptTuning}} can similarly benefit from scaling-up of the PLM, but we leave this as future work owing to resource limitations.

\begin{table}[t!]
\centering
\scalebox{0.70}{
\begin{tabular}{llccc}
\toprule
\textbf{Methods} & \textbf{Size}    & News & NER & RE \\
\toprule 
Finetune$_{\text{multitask}}$ & 770M       & 60.6$_{\pm 5.0 }$  & 64.5$_{\pm 0.2 }$   & 87.9$_{\pm 0.6 }$ \\[0.5ex]
\hline
Finetune & 770M       & 83.1$_{\pm 0.4 }$ & 77.9$_{\pm 0.4 }$  & 94.5$_{\pm 0.7 }$ \\
Finetune$_{\text{EWC}}$ & 770M       & 82.5$_{\pm 0.4 }$ & 77.3$_{\pm 0.4 }$  & 94.5$_{\pm 0.3 }$\\
Finetune$_{\text{MAS}}$ & 770M       & 59.7$_{\pm 9.8}$ & 60.4$_{\pm 3.8 }$  & 74.0$_{\pm 6.5 }$ \\
Adapter & 4.8M       & 81.4$_{\pm 0.5 }$ & 77.4$_{\pm 0.4 }$  & 94.6$_{\pm 0.5 }$ \\
Adapter$_{\text{EWC}}$ & 4.8M       & 81.5$_{\pm 0.4 }$  & 77.3$_{\pm 0.4 }$   & 94.8$_{\pm 0.8 }$  \\
Adapter$_{\text{MAS}}$ & 4.8M       & 81.0$_{\pm 0.2 }$  & 77.2$_{\pm 0.5 }$  & 94.9$_{\pm 0.6 }$   \\
PT & 0.13M       & 79.7$_{\pm 0.3 }$ & 75.4$_{\pm 0.2 }$  & 94.3$_{\pm 0.9 }$ \\
PT$_{\text{EWC}}$ & 0.13M       & 79.9$_{\pm 0.6 }$  & 74.8$_{\pm 0.6}$   & 94.5$_{\pm 0.6 }$  \\
PT$_{\text{MAS}}$ & 0.13M       & 49.6$_{\pm 4.8 }$ & 44.4$_{\pm 2.3 }$   & 84.3$_{\pm 3.8 }$   \\[0.5ex]
\hline 
ModularPT & 0.13M       & 80.4$_{\pm 0.6 }$ & 74.5$_{\pm 0.6 }$  & 93.6$_{\pm 0.4 }$  \\[0.5ex]
\bottomrule
\end{tabular}
}
\caption{\textbf{Stage-specific} performance on News Classification, NER and RE  (averaged over stages)}
\label{table:specific:all}
\end{table}

\subsection{Ablation Study} \label{sec:exp:ablation}

We now analyze the contribution of different components of \mprompt towards its SoTA performance. From the results in Table \ref{table:ablation}, we see that in stage-agnostic setting, both label prompt transfer and subset invariant loss provide a boost, though the role of the former is seemingly more significant. On the contrary, removing subset invariant loss has a more debilitating effect on stage-fused performance. This evinces that subset invariant loss is indeed critical in learning label modular representations. This is essential to the stage-fused evaluation which needs to extract and dynamically re-compose label-specific knowledge.

\begin{table}[t!]
\setlength{\tabcolsep}{2.2pt}
\centering
\scalebox{0.65}{
\begin{tabular}{lccc|ccc}
\Xhline{3\arrayrulewidth}
\multirow{2}{*}{\textbf{Methods}}    & \multicolumn{3}{c|}{Stage-agnostic} & \multicolumn{3}{c}{Stage-fused} \\ \cline{2-7}

           & News                            & NER & RE & News                            & NER & RE                                     \\
 \hline
\mprompt & 43.2$_{\pm 0.6}$ & 44.8$_{\pm 4.9}$ & 61.8$_{\pm 1.8}$ & 74.8$_{\pm 1.7}$ & 61.6$_{\pm 2.2}$ & 82.4$_{\pm 0.8}$ \\
w/o transfer & 35.4$_{\pm 2.7}$ & 43.0$_{\pm 3.6}$ & 54.8$_{\pm 2.7}$ & 64.9$_{\pm 1.7}$ & 57.4$_{\pm 2.3}$ & 77.5$_{\pm 1.2}$ \\
w/o subset-inv & 41.1$_{\pm 2.4}$ & 40.0$_{\pm 4.6}$ & 58.2$_{\pm 2.6}$ & 58.5$_{\pm 3.0}$ & 44.1$_{\pm 2.9}$ & 47.7$_{\pm 4.1}$ \\[0.5ex]
\Xhline{3\arrayrulewidth}
\end{tabular}
}
\caption{\textbf{Ablation study} of \mprompt: average performance on stage-agnostic and stage-fused settings}
\label{table:ablation}
\end{table}

\begin{table}[t!]
\centering
\scalebox{0.75}{
\begin{tabular}{lccc}
\Xhline{3\arrayrulewidth}
\textbf{Methods}    & News & NER & RE\\
\hline  
ModularPT & 74.8$_{\pm 1.7 }$ & 61.6$_{\pm 2.2 }$  & 82.4$_{\pm 0.8 }$  \\[0.5ex]
\hline
drop ground-truth label prompt & 1.5$_{\pm 0.5 }$ & 4.1$_{\pm 0.8 }$  & 0.5$_{\pm 0.4 }$ \\
drop one random label prompt & 74.9$_{\pm 1.4 }$ & 61.8$_{\pm 2.1 }$  & 83.2$_{\pm 0.6 }$ \\
permute label prompt order & 72.4$_{\pm 1.3 }$ & 61.5$_{\pm 2.3 }$  & 82.2$_{\pm 0.9 }$ \\[0.5ex]
\Xhline{3\arrayrulewidth}
\end{tabular}
}
\caption{Mean Stage-fused performance for different inference schemes}
\label{table:analysis}
\end{table}

\subsection{Quantitative Analysis} \label{sec:exp:quant_analysis}

Apart from achieving SoTA, does \mprompt possess the desirable characteristics of a modular model? According to Algorithm \ref{alg:cl}, \mprompt set $S=\SSS^{ts}_{i}$ during inference. We experiment with different strategies of input prompt construction including dropping label prompt(s) either corresponding to ground truth label(s) or one other random label, 
and permuting the default order of label prompts; see Table \ref{table:analysis} for the results. 


Indeed we observe that dropping the ground truth label prompt during inference degrades the mean performance by 57\%-82\% while dropping any other random label prompt boosts performance slightly. This strongly demonstrates the \emph{label grounding} property of \mprompt, i.e. the knowledge of a class label is exclusively embedded in its corresponding label prompt. \mprompt also shows low sensitivity to the order of label prompts during inference - a yet another favourable property of label modular models

\subsection{Qualitative Analysis} \label{sec:exp:qual_analysis}
\begin{figure}[t!]
	\centering
	\includegraphics[width=0.48\textwidth]{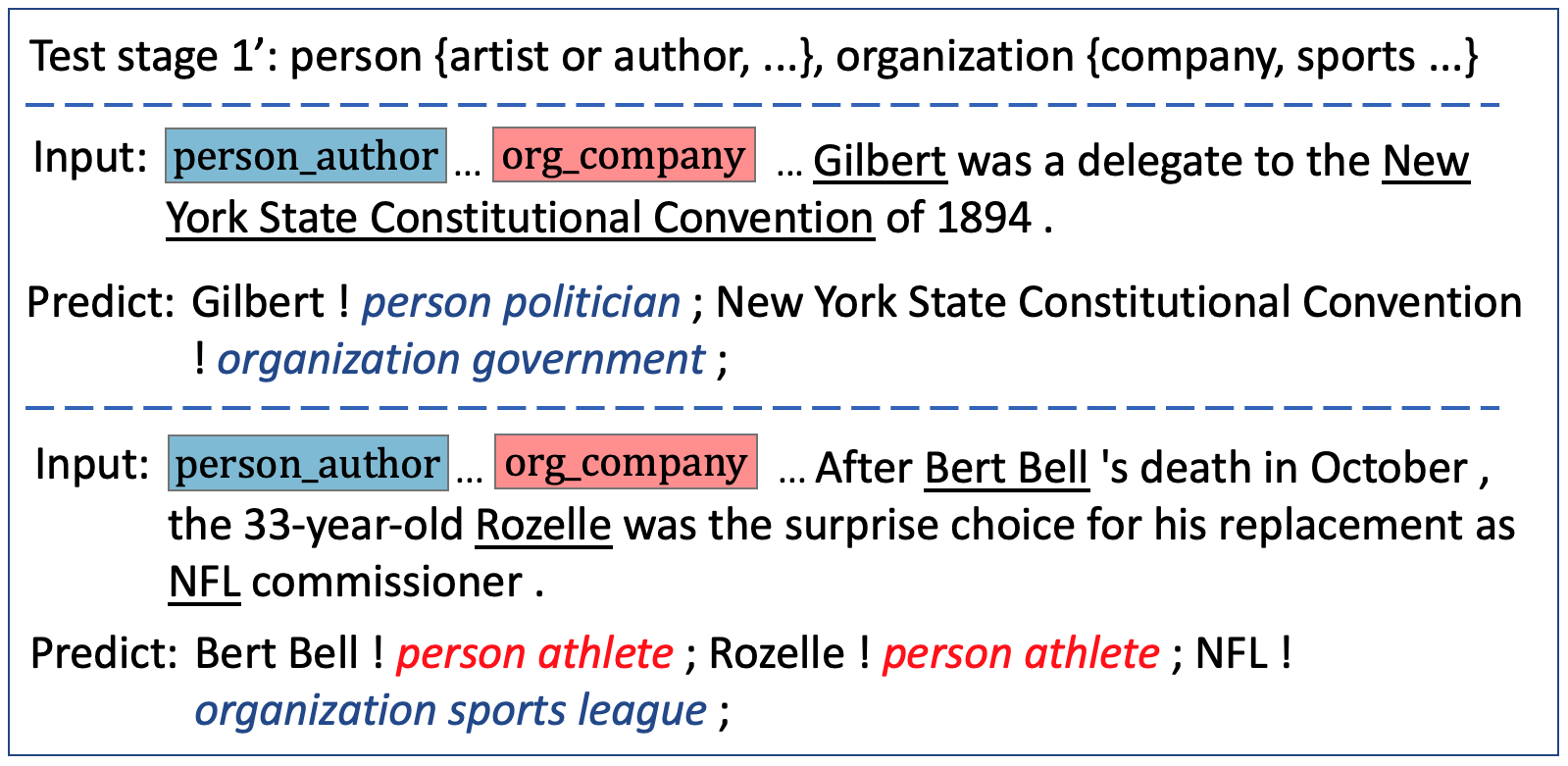}
	\caption{\darkBlue{Successful} (blue) and \red{failure} (red) cases of \mprompt predictions for stage-fused NER} 
	\label{fig:case_study_1}
\end{figure}

Revisiting Figure \ref{fig:case_study} presented in \Cref{sec:intro}, we observe that \mprompt is able to predict correctly on a testing regime that is unseen during training, by extracting and consolidating label specific knowledge from multiple training stages. More example predictions are shown in Figure \ref{fig:case_study_1} (and \cref{appendix:qual}), which indicate that \mprompt is able to exploit in-context learning over label-prompts to generalize to unseen label-combinations during inference.
For example, \mprompt tags ``Gilbert'' as politician as he was ``a delegate to'' a government. In the same spirit, \mprompt wrongly tags ``Bert Bell'' and ``Rozelle'' as athletes (true label being \emph{person\_other}) because they are associated with the sports league ``NFL''. Such qualitative findings demonstrate \mprompt's capabilities to learn label modular representations and integrate them dynamically during inference.

%% file: appendix.tex
\appendix
\section{More Details on Dataset Construction} \label{appendix:dataset}
In this section, we provide more complementary details to \Cref{sec:exp:tasks}. Table \ref{table:appendix:examples_data} shows examples of text input and target for three datasets. For Huffpost News and FewRel, we subsample exactly 200,50 examples per label class. For FewNERD, as it can have multiple label types per example, we subsample at 50 examples per label class. 

\begin{table}[H]
\centering
\scalebox{0.75}{
\begin{tabular}{p{1.5cm}p{7cm}}
\toprule
\textbf{Dataset} & \textbf{Example}  \\[0.5ex]
\toprule
\multirow{2}{*}{Huffpost} & 'Tis The Season To Be Cheeky With 'Jingle Butts' Music Video. Clip features both male and female butt models. $\rightarrow$ weird news \\ \cline{2-2}
& The 4 Best Last-Minute Christmas Gift Ideas. Scrambling to shop for that last relative, significant other or hard-to-buy-for friend? We're here to help. $\rightarrow$ style and beauty \\[0.5ex]
\hline
\multirow{2}{*}{FewNERD} & One of the Yak-3s was destroyed right away . $\rightarrow$ Yak-3s ! product airplane ;\\ \cline{2-2}
& All Together Now is a British reality television music competition which first aired on BBC One on 27 January 2018 . $\rightarrow$ All Together Now ! event other ; BBC One ! organization media ;\\[0.5ex]
\hline
\multirow{2}{*}{FewRel} & Its main base was at Tampere - Pirkkala Airport ( TMP ) , Tampere . Pirkkala Airport | Tampere $\rightarrow$ place served by trainsport hub\\ \cline{2-2}
& He represented Sweden at the 1934 FIFA World Cup . 1934 FIFA World Cup | Sweden $\rightarrow$ participating teams\\[0.5ex]
\bottomrule
\end{tabular}
}
\caption{Examples of text input $\rightarrow$ text target for Huffpost News, fewNERD and FewRel}
\label{table:appendix:examples_data}
\end{table}

\section{More Qualitative Examples} \label{appendix:qual}
\begin{figure}[H]
	\centering
	\includegraphics[width=0.48\textwidth]{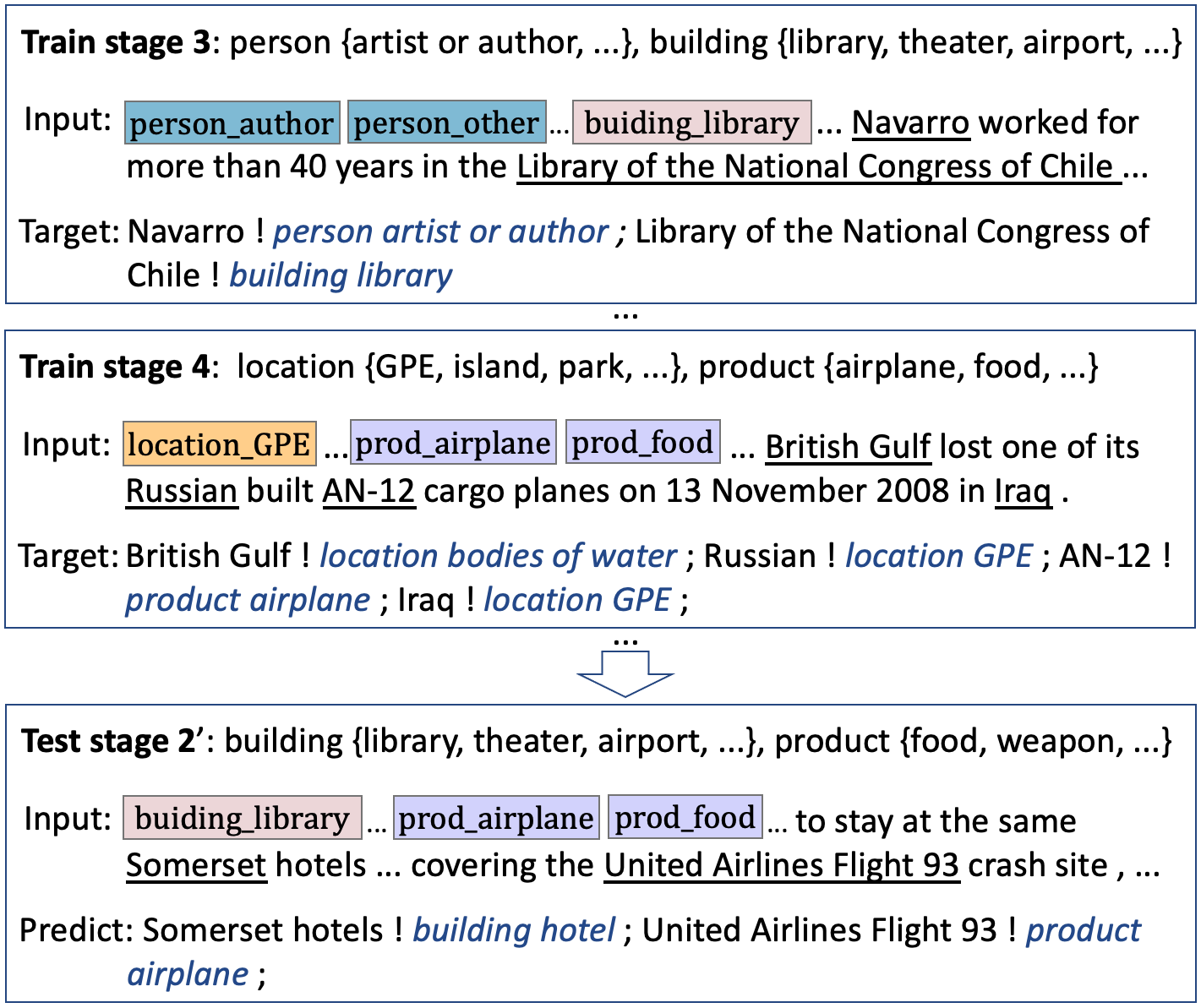}
	\caption{\small Another example of \mprompt for stage-fused NER} 
	\label{fig:appendix:case_study}
\end{figure}

\begin{figure}[H]
	\centering
	\includegraphics[width=0.48\textwidth]{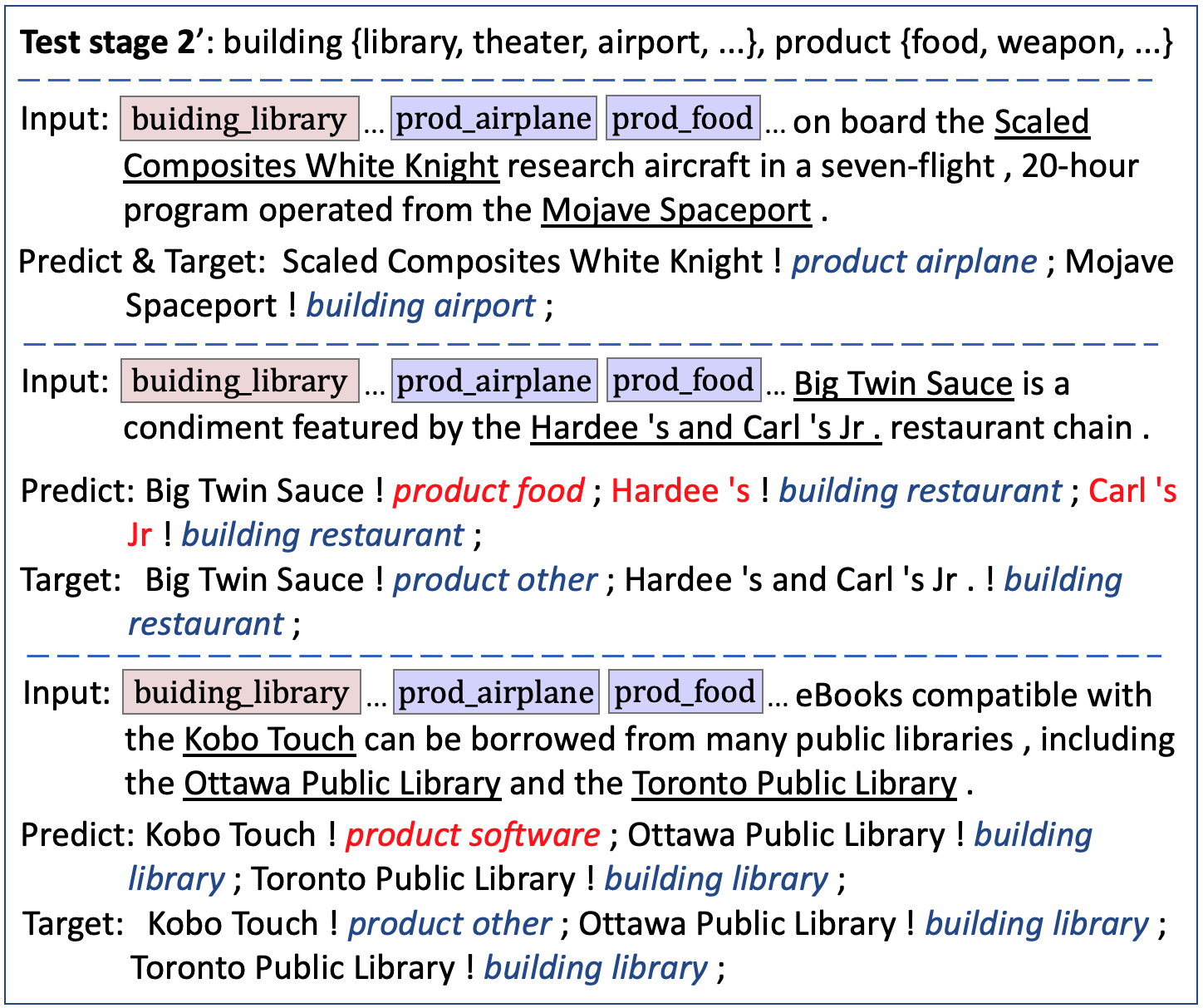}
	\caption{\darkBlue{Successful} (blue) and \red{failure} (red) cases of \mprompt predictions for stage-fused NER} 
	\label{fig:appendix:case_study_1}
\end{figure}

Similar to \Cref{sec:exp:qual_analysis}, we show another example of \mprompt on stage-fused NER in Figure \ref{fig:appendix:case_study} and more example predictions in Figure \ref{fig:appendix:case_study_1}. These additional examples strengthen the conclusions in \Cref{sec:exp:qual_analysis}. In the second example in Figure \ref{fig:appendix:case_study_1}, \mprompt tags "Big Twin Sauce" as a \textit{product food} while its ground truth tag is \textit{product other}. We can see \mprompt considers the context as the entity is associated with a restaurant. Similarly, in the third example, "Kobo Touch" is actually a hardware reader and its ground truth tag is \textit{product other}. However, such world knowledge is not available and \mprompt tags it as a software based on the context of "eBooks" and libraries.

\section{Detailed Results on Stage-agnostic and Stage-specific Settings} \label{sec:appendix_results}

Table \ref{appendix:table:agnostic:all} shows detailed stage-agnostic results. $D_{\{1..k\}}$ corresponds to multi-stage training over domains $\{1,..,k\}$ and testing over the aggregated domains. For baselines, we use their checkpoints after k-th stage for evaluating $D_{\{1..k\}}$. For \mprompt, we use the final checkpoint. Table \ref{appendix:fuse:news_ner}, \ref{appendix:fuse:re} show detailed stage-fused results. $D_k$' denotes k-th task fused stage.

\begin{table*}[htbp!]
\centering
\scalebox{0.65}{
\begin{tabular}{lllll|lll|llll}
\Xhline{3\arrayrulewidth}
\multirow{2}{*}{\textbf{Methods}}    & \multicolumn{4}{c|}{News Classification} & \multicolumn{3}{c|}{NER} & \multicolumn{4}{c}{RE}\\ \cline{2-6} \cline{7-10} \cline{11-12}

                                       & D$_{\{1..2\}}$                         & D$_{\{1..3\}}$                       & D$_{\{1..4\}}$                       & D$_{\{1..5\}}$                               & D$_{\{1..2\}}$                         & D$_{\{1..3\}}$                       & D$_{\{1..4\}}$                             & D$_{\{1..2\}}$                         & D$_{\{1..3\}}$                       & D$_{\{1..4\}}$                       & D$_{\{1..5\}}$       \\
\hline  
Finetune$_{Multitask}$   &  60.0$_{\pm 5.2 }$  &  61.6$_{\pm 4.8 }$  &  62.1$_{\pm 4.3 }$  &  60.6$_{\pm 5.0 }$  &  64.3$_{\pm 0.3 }$     &  68.1$_{\pm 0.3 }$       &  64.5$_{\pm 0.2 }$  &  85.3$_{\pm 1.0 }$  &  87.4$_{\pm 0.6 }$  &  83.7$_{\pm 2.2 }$  &  87.9$_{\pm 0.6 }$\\[1ex]
\hline
Finetune    &  57.8$_{\pm 0.9 }$           &  35.2$_{\pm 1.8 }$           &  29.4$_{\pm 1.2 }$           &  23.0$_{\pm 0.6 }$  &  39.8$_{\pm 0.4 }$           &  30.9$_{\pm 2.1 }$  &  25.0$_{\pm 2.7 }$  &  71.1$_{\pm 2.2 }$           &  57.4$_{\pm 2.5 }$           &  48.0$_{\pm 1.6 }$           &  40.8$_{\pm 2.2 }$\\ 

Finetune$_{EWC}$  &  \textbf{61.4}$_{\pm 2.6 }$           &  35.3$_{\pm 1.8 }$           &  27.3$_{\pm 2.5 }$           &  22.5$_{\pm 1.2 }$  &  42.9$_{\pm 2.2 }$           &  37.8$_{\pm 4.8 }$  &  26.6$_{\pm 2.8 }$  &  79.2$_{\pm 3.5 }$           &  61.2$_{\pm 6.5 }$           &  52.5$_{\pm 3.4 }$           &  45.4$_{\pm 3.2 }$ \\

Finetune$_{MAS}$  &  58.9$_{\pm 1.1}$ & 46.9$_{\pm 2.8}$ & 38.7$_{\pm 3.0}$ & 29.2$_{\pm 2.9}$  &  48.6$_{\pm 7.5 }$  &  37.1$_{\pm 2.1 }$  &  25.4$_{\pm 2.9 }$  &  74.5$_{\pm 3.9 }$  &  61.3$_{\pm 2.7 }$  &  54.7$_{\pm 1.5 }$  &  46.2$_{\pm 1.8 }$\\

Adapter  &  45.2$_{\pm 2.6 }$           &  29.3$_{\pm 0.4 }$           &  24.5$_{\pm 0.5 }$           &  19.3$_{\pm 0.3 }$  &  38.9$_{\pm 1.0 }$  &  27.7$_{\pm 0.7 }$  &  21.2$_{\pm 2.0 }$  &  56.6$_{\pm 1.4 }$           &  40.4$_{\pm 2.5 }$           &  36.9$_{\pm 1.2 }$           &  28.6$_{\pm 1.9 }$\\

Adapter$_{EWC}$  &  45.6$_{\pm 3.0}$ & 29.6$_{\pm 0.5}$ & 24.4$_{\pm 0.6}$ & 18.6$_{\pm 1.0}$  &  38.6$_{\pm 0.7}$ & 29.1$_{\pm 1.7}$ & 20.5$_{\pm 1.3}$  &   54.9$_{\pm 1.2}$ & 42.6$_{\pm 1.5}$ & 36.1$_{\pm 0.8}$ & 27.8$_{\pm 1.2}$  \\

Adapter$_{MAS}$  &  45.6$_{\pm 2.6}$  &  30.0$_{\pm 0.8}$  &  26.1$_{\pm 2.5}$  &  20.1$_{\pm 1.5}$  &  38.2$_{\pm 0.3}$  &  28.6$_{\pm 1.1}$  &  21.1$_{\pm 2.0}$  &  57.1$_{\pm 2.9}$  &  40.5$_{\pm 1.9}$  &  35.5$_{\pm 1.9}$  &  28.3$_{\pm 0.5}$ \\

PT  &  36.9$_{\pm 0.5}$  &  27.9$_{\pm 0.4}$  &  21.7$_{\pm 0.4}$  &  16.1$_{\pm 0.3}$  &  37.0$_{\pm 0.3 }$           &  26.1$_{\pm 0.3 }$   &  18.8$_{\pm 0.1 }$  &  47.3$_{\pm 1.0 }$  &  32.4$_{\pm 0.5 }$  &  24.3$_{\pm 0.4 }$  &  19.6$_{\pm 0.1 }$ \\

PT$_{EWC}$  &  36.9$_{\pm 1.1}$  &  28.2$_{\pm 0.2}$  &  21.5$_{\pm 0.3}$  &  16.5$_{\pm 0.5}$   &  36.7$_{\pm 0.8}$  &  26.0$_{\pm 0.4}$  &  19.0$_{\pm 0.5}$   &  46.8$_{\pm 0.5}$ & 32.3$_{\pm 1.2}$ & 25.4$_{\pm 0.9}$ & 20.1$_{\pm 0.8}$ \\

PT$_{MAS}$  &  43.0$_{\pm 3.5}$  &  30.1$_{\pm 5.2}$  &  24.1$_{\pm 4.5}$  &  23.8$_{\pm 5.9}$  &  28.8$_{\pm 2.8}$  &  29.2$_{\pm 6.0}$  &  29.3$_{\pm 4.6}$  &  51.3$_{\pm 1.3}$  &  36.0$_{\pm 0.9}$  &  30.7$_{\pm 1.7}$  &  25.1$_{\pm 4.9}$  \\

PT$_{CL}$  &  60.3$_{\pm 3.7 }$  &  41.8$_{\pm 7.5 }$  &  35.4$_{\pm 3.3 }$  &  28.1$_{\pm 2.9 }$  &  47.4$_{\pm 4.6 }$  &  39.7$_{\pm 3.9 }$  &  30.9$_{\pm 5.4 }$  &  69.9$_{\pm 7.8 }$  &  54.3$_{\pm 3.9 }$  &  47.9$_{\pm 4.2 }$  &  43.1$_{\pm 4.9 }$\\[1ex]
\hline
ModularPT        &  60.3$_{\pm 3.9 }$  &  \textbf{53.9}$_{\pm 2.3 }$  &  \textbf{51.1}$_{\pm 1.4 }$  &  \textbf{43.2}$_{\pm 0.6 }$  &  \textbf{63.3}$_{\pm 1.4 }$  &  \textbf{53.9}$_{\pm 4.3 }$  &  \textbf{44.8}$_{\pm 4.9 }$  &  \textbf{81.3}$_{\pm 3.3 }$  &  \textbf{70.1}$_{\pm 3.7 }$  &  \textbf{65.1}$_{\pm 2.2 }$  &  \textbf{61.8}$_{\pm 1.8 }$\\
w/o transfer & 61.2$_{\pm 2.8}$ & 48.9$_{\pm 2.4}$ & 43.1$_{\pm 2.7}$ & 35.4$_{\pm 2.7}$ & 62.6$_{\pm 1.4}$ & 53.8$_{\pm 1.7}$ & 43.0$_{\pm 3.6}$ & 82.2$_{\pm 2.3}$ & 67.2$_{\pm 3.2}$ & 59.4$_{\pm 2.0}$ & 54.8$_{\pm 2.7}$\\
w/o subset\_inv & 62.4$_{\pm 1.9}$ & 53.9$_{\pm 3.2}$ & 50.0$_{\pm 2.9}$ & 41.1$_{\pm 2.4}$ & 60.0$_{\pm 2.8}$ & 51.2$_{\pm 3.5}$ & 40.0$_{\pm 4.6}$ & 74.4$_{\pm 10.9}$ & 66.2$_{\pm 4.8}$ & 60.3$_{\pm 3.5}$ & 58.2$_{\pm 2.6}$\\[1ex]
\Xhline{3\arrayrulewidth}
\end{tabular}
}
\caption{Detailed Stage-agnostic Performance on News Classification, NER and RE.}
\label{appendix:table:agnostic:all}
\end{table*}

\begin{table*}
\centering
\scalebox{0.67}{
\begin{tabular}{lllllll|lllll}
\Xhline{3\arrayrulewidth}
\multirow{2}{*}{\textbf{Methods}}    & \multicolumn{6}{c|}{News Classification} & \multicolumn{5}{c}{NER} \\ \cline{2-7} \cline{8-12} 

           & D1'                            & D2'                     & D3'                       & D4'                       & D5' & Mean   & D1'                            & D2'                     & D3'                       & D4' & Mean  \\
\hline  
Finetune$_{Multitask}$   &  58.5$_{\pm 0.7 }$  &  57.5$_{\pm 1.7 }$  &  60.8$_{\pm 2.7 }$  &  64.4$_{\pm 1.8 }$  &  62.2$_{\pm 1.6 }$  &  60.7$_{\pm 0.6 }$  &  64.3$_{\pm 1.0 }$  &  53.8$_{\pm 0.1 }$  &  51.0$_{\pm 0.7 }$  &  59.1$_{\pm 0.2 }$  &  57.0$_{\pm 0.0 }$  \\ [0.5ex]
\hline 
Finetune    &  20.5$_{\pm 1.5 }$           &  21.7$_{\pm 1.2 }$           &  21.8$_{\pm 1.3 }$           &  24.5$_{\pm 1.3 }$           &  23.4$_{\pm 1.0 }$  &  22.4$_{\pm 0.5 }$  &  20.9$_{\pm 2.9 }$  &  24.1$_{\pm 2.9 }$  &  35.4$_{\pm 1.8 }$  &  19.1$_{\pm 3.4 }$  &  24.9$_{\pm 2.1 }$   \\
Finetune$_{EWC}$  &  20.9$_{\pm 1.7 }$  &  21.9$_{\pm 0.7 }$  &  21.2$_{\pm 2.2 }$  &  25.6$_{\pm 1.8 }$  &  23.3$_{\pm 1.7 }$  &  22.6$_{\pm 1.0 }$  &  21.8$_{\pm 1.5 }$  &  24.6$_{\pm 1.4 }$  &  34.7$_{\pm 2.0 }$  &  19.3$_{\pm 4.2 }$  &  25.1$_{\pm 1.7 }$  \\
Finetune$_{MAS}$  &  26.4$_{\pm 6.2 }$  &  25.6$_{\pm 2.9 }$  &  26.8$_{\pm 3.4 }$  &  38.4$_{\pm 7.0 }$  &  26.5$_{\pm 6.4 }$  &  28.8$_{\pm 3.3 }$  &  22.4$_{\pm 3.5 }$  &  25.2$_{\pm 0.9 }$  &  19.1$_{\pm 4.4 }$  &  27.2$_{\pm 5.9 }$  &  23.5$_{\pm 1.3 }$  \\
Adapter  &  15.8$_{\pm 0.9 }$            &  20.5$_{\pm 0.4 }$           &  15.5$_{\pm 2.0 }$                &  22.6$_{\pm 0.6 }$           &  19.6$_{\pm 1.2 }$  &  18.8$_{\pm 0.3 }$   &  17.1$_{\pm 2.0 }$  &  21.6$_{\pm 3.3 }$  &  33.5$_{\pm 1.7 }$  &  15.0$_{\pm 1.9 }$  &  21.8$_{\pm 1.7 }$  \\
Adapter$_{EWC}$  &  15.2$_{\pm 1.3}$ & 20.4$_{\pm 0.5}$ & 13.7$_{\pm 2.0}$ & 23.5$_{\pm 1.9}$ & 18.0$_{\pm 1.4}$ & 18.1$_{\pm 1.0}$   &  17.3$_{\pm 1.8}$ & 21.0$_{\pm 2.5}$ & 33.8$_{\pm 1.6}$ & 14.5$_{\pm 1.5}$ & 21.7$_{\pm 1.5}$  \\
Adapter$_{MAS}$  &  17.3$_{\pm 2.0}$ & 20.2$_{\pm 0.4}$ & 16.5$_{\pm 1.3}$ & 23.5$_{\pm 2.3}$ & 20.3$_{\pm 2.5}$ & 19.6$_{\pm 1.4}$   &  16.7$_{\pm 2.1}$ & 20.4$_{\pm 2.5}$ & 34.1$_{\pm 2.3}$ & 14.2$_{\pm 2.0}$ & 21.4$_{\pm 2.0}$   \\
PT  &  13.7$_{\pm 0.3 }$            &  19.6$_{\pm 0.3 }$           &  8.3$_{\pm 0.2 }$                &  21.0$_{\pm 1.3 }$           &  15.6$_{\pm 0.6 }$  &  15.7$_{\pm 0.2 }$   &  14.9$_{\pm 1.2 }$  &  17.2$_{\pm 0.8 }$  &  31.9$_{\pm 1.3 }$  &  13.2$_{\pm 1.0 }$  &  19.3$_{\pm 0.5 }$  \\
PT$_{EWC}$ & 13.7$_{\pm 0.4}$ & 19.7$_{\pm 0.2}$ & 9.2$_{\pm 1.5}$ & 21.9$_{\pm 0.5}$ & 15.8$_{\pm 1.1}$ & 16.1$_{\pm 0.5}$ & 14.7$_{\pm 0.8}$ & 18.5$_{\pm 1.9}$ & 33.0$_{\pm 1.3}$ & 14.0$_{\pm 1.5}$ & 20.0$_{\pm 1.0}$ \\
PT$_{MAS}$  &  25.3$_{\pm 6.0}$ & 21.9$_{\pm 3.9}$ & 20.0$_{\pm 13.4}$ & 27.6$_{\pm 7.9}$ & 23.8$_{\pm 5.3}$ & 23.7$_{\pm 6.2}$   &  26.1$_{\pm 3.1}$ & 20.7$_{\pm 2.5}$ & 22.3$_{\pm 4.2}$ & 31.9$_{\pm 7.8}$ & 25.3$_{\pm 3.8}$  \\
PT$_{CL}$  &  25.7$_{\pm 5.9 }$  &  27.1$_{\pm 5.5 }$  &  21.9$_{\pm 5.5 }$  &  29.1$_{\pm 7.3 }$  &  33.9$_{\pm 4.8 }$  &  27.5$_{\pm 2.8 }$   &  28.1$_{\pm 4.3 }$  &  34.7$_{\pm 4.8 }$  &  30.5$_{\pm 6.0 }$  &  28.2$_{\pm 7.9 }$  &  30.4$_{\pm 5.1 }$  \\[0.5ex]
\hline
Finetune$^*_{MAS}$ & 47.9$_{\pm 2.4}$ & 49.7$_{\pm 5.5}$ & 55.8$_{\pm 4.7}$ & 52.4$_{\pm 2.5}$ & 58.2$_{\pm 4.9}$ & 52.8$_{\pm 2.2}$  & 23.3$_{\pm 2.8}$ & 29.3$_{\pm 1.8}$ & 22.3$_{\pm 1.7}$ & 28.3$_{\pm 2.0}$ & 25.8$_{\pm 1.7}$ \\
PT$^*_{CL}$ & 56.1$_{\pm 6.3}$ & 62.5$_{\pm 4.7}$ & 57.5$_{\pm 6.6}$ & 55.8$_{\pm 9.4}$ & 55.5$_{\pm 2.0}$ & 57.5$_{\pm 5.2}$  & 33.1$_{\pm 4.0}$ & 37.4$_{\pm 1.4}$ & 28.2$_{\pm 4.9}$ & 38.6$_{\pm 2.0}$ & 34.3$_{\pm 2.4}$ \\[0.5ex]
\hline 
ModularPT        &  \textbf{71.6}$_{\pm 3.0 }$  &  \textbf{75.8}$_{\pm 4.0 }$  &  \textbf{76.5}$_{\pm 2.0 }$  &  \textbf{79.1}$_{\pm 3.3 }$  &  \textbf{71.0}$_{\pm 1.6 }$  &  \textbf{74.8}$_{\pm 1.7 }$  &  \textbf{62.8}$_{\pm 4.0 }$  &  \textbf{60.7}$_{\pm 1.1 }$  &  \textbf{61.5}$_{\pm 2.5 }$  &  \textbf{61.5}$_{\pm 4.4 }$  &  \textbf{61.6}$_{\pm 2.2 }$  \\
w/o transfer & 67.8$_{\pm 3.9}$ & 69.7$_{\pm 3.0}$ & 66.7$_{\pm 5.3}$ & 63.8$_{\pm 3.3}$ & 56.6$_{\pm 5.3}$ & 64.9$_{\pm 1.7}$ &  56.0$_{\pm 4.5}$ & 51.2$_{\pm 7.6}$ & 61.3$_{\pm 1.6}$ & 61.2$_{\pm 2.7}$ & 57.4$_{\pm 2.3}$ \\
w/o subset\_inv & 61.3$_{\pm 7.7}$ & 55.9$_{\pm 11.8}$ & 60.4$_{\pm 5.2}$ & 61.0$_{\pm 3.0}$ & 53.7$_{\pm 7.2}$ & 58.5$_{\pm 3.0}$ &  51.9$_{\pm 9.5}$ & 37.5$_{\pm 3.7}$ & 40.7$_{\pm 7.3}$ & 46.3$_{\pm 4.3}$ & 44.1$_{\pm 2.9}$\\
[0.5ex]
\Xhline{3\arrayrulewidth}
\end{tabular}
}
\caption{Detailed Stage-fused Performance on News Classification and NER}
\label{appendix:fuse:news_ner}
\end{table*}

\begin{table}[H]
\centering
\scalebox{0.58}{
\begin{tabular}{lllllll}
\Xhline{3\arrayrulewidth}
\multirow{2}{*}{\textbf{Methods}}    & \multicolumn{5}{c}{RE} &\multirow{2}{*}{\textbf{Mean}}\\ \cline{2-6}

           & D1'                            & D2'                            & D3'                            & D4'                            & D5'          \\
\hline  
Finetune$_{Multitask}$   &  80.5$_{\pm 2.1 }$  &  90.3$_{\pm 1.6 }$  &  85.4$_{\pm 1.2 }$  &  88.5$_{\pm 0.9 }$  &  85.4$_{\pm 1.0 }$  &  86.0$_{\pm 0.9 }$\\[1ex]
\hline
Finetune    &  31.2$_{\pm 2.8 }$  &  49.6$_{\pm 3.5 }$  &  37.4$_{\pm 4.0 }$  &  43.3$_{\pm 3.7 }$  &  42.2$_{\pm 1.3 }$  &  40.7$_{\pm 2.1 }$\\ 

Finetune$_{EWC}$  &  34.3$_{\pm 4.1 }$  &  57.8$_{\pm 3.6 }$  &  40.6$_{\pm 2.5 }$  &  46.9$_{\pm 6.0 }$  &  46.5$_{\pm 2.1 }$  &  45.2$_{\pm 3.3 }$\\

Finetune$_{MAS}$  &  41.7$_{\pm 3.0 }$  &  60.1$_{\pm 3.4 }$  &  52.2$_{\pm 4.2 }$  &  46.1$_{\pm 3.8 }$  &  34.4$_{\pm 3.7 }$  &  46.9$_{\pm 1.9 }$\\

Adapter  &  21.0$_{\pm 0.6 }$  &  33.8$_{\pm 4.6 }$  &  21.3$_{\pm 1.3 }$  &  32.9$_{\pm 2.0 }$  &  32.0$_{\pm 2.6 }$  &  28.2$_{\pm 1.8 }$\\

Adapter$_{EWC}$  &  20.1$_{\pm 0.7}$ & 31.1$_{\pm 3.7}$ & 19.7$_{\pm 0.3}$ & 34.0$_{\pm 2.8}$ & 32.0$_{\pm 3.4}$ & 27.4$_{\pm 1.2}$  \\

Adapter$_{MAS}$ & 20.1$_{\pm 0.5}$ & 32.2$_{\pm 2.1}$ & 22.5$_{\pm 1.6}$ & 31.6$_{\pm 3.1}$ & 32.7$_{\pm 2.4}$ & 27.8$_{\pm 0.6}$ \\

PT  &  19.6$_{\pm 0.1 }$  &  20.8$_{\pm 0.7 }$  &  20.3$_{\pm 1.2 }$  &  18.5$_{\pm 0.2 }$  &  19.0$_{\pm 0.3 }$  &  19.6$_{\pm 0.2 }$ \\

PT$_{EWC}$  &  19.8$_{\pm 0.1}$ & 22.2$_{\pm 3.0}$ & 19.9$_{\pm 0.4}$ & 18.5$_{\pm 0.1}$ & 19.8$_{\pm 1.1}$ & 20.0$_{\pm 0.8}$ \\

PT$_{MAS}$ & 18.8$_{\pm 6.9}$ & 25.0$_{\pm 9.1}$ & 22.8$_{\pm 5.9}$ & 32.7$_{\pm 8.1}$ & 25.5$_{\pm 3.7}$ & 24.9$_{\pm 5.1}$ \\

PT$_{CL}$  &  37.6$_{\pm 9.1 }$  &  51.1$_{\pm 4.4 }$  &  37.8$_{\pm 4.4 }$  &  41.4$_{\pm 4.7 }$  &  48.4$_{\pm 6.1 }$  &  43.3$_{\pm 5.2 }$\\[0.5ex]
\hline
Finetune$^*_{MAS}$  &  55.7$_{\pm 3.8}$ & 72.9$_{\pm 2.6}$ & 69.8$_{\pm 3.5}$ & 80.2$_{\pm 2.2}$ & 62.0$_{\pm 2.8}$ & 68.1$_{\pm 1.3}$\\
PT$^*_{CL}$  &  58.8$_{\pm 5.1}$ & 65.1$_{\pm 4.0}$ & 61.1$_{\pm 1.6}$ & 78.3$_{\pm 3.3}$ & 63.3$_{\pm 3.5}$ & 65.3$_{\pm 2.7}$\\[0.5ex]
\hline
ModularPT        &  \textbf{80.3}$_{\pm 1.6 }$  &  \textbf{89.2}$_{\pm 2.5 }$  &  \textbf{75.1}$_{\pm 3.9 }$  &  \textbf{88.5}$_{\pm 2.7 }$  &  \textbf{78.6}$_{\pm 0.8 }$  &  \textbf{82.4}$_{\pm 0.8 }$ \\
w/o transfer & 73.5$_{\pm 2.6}$ & 85.1$_{\pm 1.9}$ & 66.9$_{\pm 7.4}$ & 88.2$_{\pm 2.6}$ & 73.7$_{\pm 3.6}$ & 77.5$_{\pm 1.2}$\\
w/o subset\_inv & 45.1$_{\pm 5.1}$ & 50.6$_{\pm 6.1}$ & 48.3$_{\pm 5.9}$ & 46.5$_{\pm 4.8}$ & 48.0$_{\pm 6.7}$ & 47.7$_{\pm 4.1}$\\
[1ex]
\Xhline{3\arrayrulewidth}
\end{tabular}
}
\caption{Detailed Stage-fused Performance on RE}
\label{appendix:fuse:re}
\end{table}